\pdfoutput=1

\documentclass[11pt]{article}

\usepackage[preprint]{acl}

\usepackage{times}
\usepackage{latexsym}

\usepackage[T1]{fontenc}

\usepackage[utf8]{inputenc}

\usepackage{microtype}

\usepackage{inconsolata}

\usepackage{graphicx}
\usepackage{algorithm}
\usepackage{algorithmic}
\usepackage{booktabs}
\usepackage{amsmath}
\usepackage{booktabs}
\usepackage{bbm}
\usepackage{subcaption}
\usepackage{multicol}
\usepackage{makecell}
\usepackage{multirow}
\usepackage{listings}
\usepackage{pifont}
\usepackage{xcolor}

\usepackage{microtype}
\usepackage{amssymb}

\title{Re-Thinking the Automatic Evaluation of Image-Text Alignment in Text-to-Image Models}

\author{Huixuan Zhang, Xiaojun Wan \\
  Wangxuan Institute of Computer Technology, Peking University \\
  \texttt{zhanghuixuan@stu.pku.edu.cn}, 
  \texttt{wanxiaojun@pku.edu.cn} \\}

\begin{document}
\maketitle
\begin{abstract}
Text-to-image models often struggle to generate images that precisely match textual prompts. Prior research has extensively studied the evaluation of image-text alignment in text-to-image generation. However, existing evaluations primarily focus on agreement with human assessments, neglecting other critical properties of a trustworthy evaluation framework. In this work, we first identify two key aspects that a reliable evaluation should address. We then empirically demonstrate that current mainstream evaluation frameworks fail to fully satisfy these properties across a diverse range of metrics and models. Finally, we propose recommendations for improving image-text alignment evaluation.

\end{abstract}

\section{Introduction}

Text-to-Image (T2I) models have demonstrated remarkable capabilities in generating high-quality, realistic images \cite{betker2023improving, esser2024scaling}. Despite these advances, they still face challenges in accurately interpreting and adhering to user prompts. Common failures include generating incorrect object counts, attributes, or spatial relationships \citep{Li_2024_CVPR}. Nevertheless, evaluating text-image alignment remains a persistent and unresolved problem in the field.

Several frameworks exist for evaluating image-text alignment. Model-based approaches include CLIPScore \citep{hessel2021clipscore} and VQAScore \citep{lin2024evaluating}. Component-based methods decompose text prompts into fine-grained elements and assess alignment through techniques like question generation and answering (QG/A) \cite{Hu_2023_ICCV, cho2023davidsonian}. Additionally, detector-based frameworks such as UniDet-Eval \cite{t2icompbench} leverage object detection for evaluation.

\begin{figure}[htbp]
    \centering
    \includegraphics[width = 0.48\textwidth]{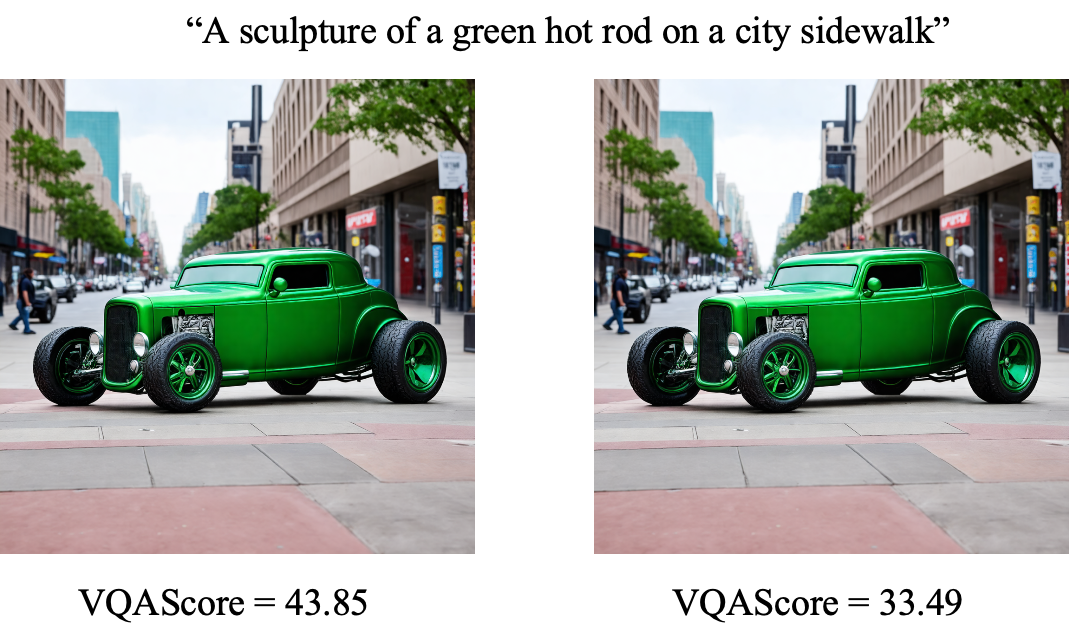}
    \caption{A robustness failure case of VQAScore. The right image is visually the same as the left one, yet their calculated VQAScore differs a lot. }
    \label{fig:intro}
    \vspace{-10pt}
\end{figure}

Despite the vast number of evaluation frameworks, few studies have thoroughly investigated the trustworthiness of image-text alignment assessment. Most existing work focuses narrowly on aligning automatic evaluation results with human judgments \cite{Li_2024_CVPR, gecko}. When addressing trustworthiness, researchers primarily concentrate on improving human evaluation protocols to better validate automatic metrics \cite{otani2023toward, gecko}. However, these approaches overlook the critical point that a truly trustworthy evaluation framework must encompass more dimensions than mere correlation with human assessment. Additional discussion of related works is provided in Appendix \ref{sec:appendix}.

In this work, we identify two critical properties for trustworthy evaluation frameworks: \textbf{Robustness} and \textbf{Significance}. Through empirical analysis, we demonstrate that current evaluation methods fail to fully satisfy these criteria, highlighting an important research direction for improving text-image alignment assessment. For example, as illustrated in Figure \ref{fig:intro}, while two visually similar images receive substantially different VQAScore evaluations, revealing clear deficiencies in Robustness.

In summary, our work makes two key contributions: (1) identifying Robustness and Significance as fundamental requirements for trustworthy evaluation frameworks, and (2) systematically demonstrating how current frameworks fail to meet these criteria.

\section{Methodology}

\subsection{Preliminaries}
We would like to briefly introduce the common image-text alignment evaluation framework that we discuss in this research. Generally, given a benchmark consisting a set of $N$ prompts $P = \{p_1, ..., p_N\}$, a text-to-image model $M$ is used to generate images $I = \{I_1,...,I_N\}$ based on these prompts. \footnote{Though there can be multiple images generated using a certain prompt, many practices only generate one image per prompt.} An automatic metric $J$ is used to evaluate the alignment between the image-text pair $(p_{i}, I_{i})$, and outputs a score $s_{i} = J(p_{i}, I_{i})$. The final evaluation result is the average of all scores: $s_{M} = \dfrac{1}{N}\sum_{i=1}^{N}s_{i}$. If there are multiple models $M_{1},...,M_{K}$, the scores $s_{M_{1}},..., s_{M_{K}}$ also provides a ranking of these models, which is also a part of the evaluation result that people focus on. 

\subsection{Aspects}
We would like to propose two important aspects that a trustworthy image-text alignment evaluation should focus on. 

\paragraph{Robustness}
Robustness requires that evaluation results remain consistent under reasonable perturbations of the input pair $(p_i, I_i)$. In this work, we specifically examine two critical dimensions of robustness: (1) robustness to randomness and (2) robustness to image perturbations.

\subparagraph{Robustness to Randomness}
Most state-of-the-art text-to-image models employ denoising diffusion processes, where the generation output for a given prompt $p_i$ depends on the sampled noise prior. This inherent randomness introduces variability in evaluation results. A trustworthy evaluation framework must maintain consistent model rankings despite this randomness - otherwise, the evaluation fails to reliably indicate which model performs better. To assess robustness under randomness, we systematically evaluate model performance across different random seeds.

\subparagraph{Robustness to Image Perturbation} 
\label{sec:2}
We make a fundamental assumption that for visually similar images $I_{i}$ and $I_{i}^{'}$, their evaluation scores $J(p_i, I_i)$ and $J(p_i, I_i^{'})$ should also be close. Large discrepancies would indicate potential metric flaw rather than true model capability. To assess robustness against image perturbations, we apply a minimal transformation: given an image $I$ with pixel values $I_{c, h, w} \in [0, 255], c, h, w$ corresponds to the channels, height and width of the image, we achieve a perturbed image $I^{'}$ with pixel values \footnote{We confirm the perturbation's visual imperceptibility through manual inspection of multiple cases, verifying that modified images remain indistinguishable from originals.}:
\begin{equation}
I^{'}_{c,h,w} = \begin{cases}
I_{c,h,w} + 1 & \text{if } I_{c,h,w} < 255 \\
I_{c,h,w} & \text{otherwise}
\end{cases}
\end{equation} 
The robustness metric is computed by calculating the performance gap as: 

\begin{equation}
\Delta J_{i} = |J(p_i, I_i) - J(p_i, I^{'}_{i})|
\end{equation}


\paragraph{Significance}
The property of Significance examines whether an observed performance difference (e.g., $s_{M_{1}} > s_{M_{2}}$) reflects a meaningful superiority of model $M_{1}$ over $M_{2}$. To quantify this, we employ two complementary approaches: 
\begin{enumerate}
    \item Statistical Testing: For evaluation score sets $S_{M_{1}}=(s_{M_{1}}^{1},...,s_{M_{1}}^{N})$ and $S_{M_{2}}=(s_{M_{2}}^{1},...,s_{M_{2}}^{N})$, we conduct a paired t-test to assess statistical significance.
    \item Dominance Ratio: We compute the empirical probability of $M_{1}$ over $M_{2}$ as:
    \begin{equation}
        R = \dfrac{1}{N}\sum_{i=1}^{N}\mathbb{I}[s_{M_{1}}^{i} > s_{M_{2}}^{i}]
    \end{equation}
\end{enumerate}

\begin{table*}[htbp]
\small
\setlength{\tabcolsep}{1.8mm}
    \begin{tabular}{c|ccc|ccc|ccc}
    \toprule
        \multirow{2}{*}{Model Name} & \multicolumn{3}{c}{VQAScore} & \multicolumn{3}{c}{CLIPScore} & \multicolumn{3}{c}{DSGScore} \\
         \cmidrule{2-10}
         ~ & 42 & 3407 & 5096 & 42 & 3407 & 5096 & 42 & 3407 & 5096  \\
         \midrule
         Stable-Diffusion-3 & 91.18(1) & 90.90(1) & 91.06(1) & 26.39(1) & 26.34(1) & 26.36(1) & 93.66(1) & 91.99(1) & 93.52(1) \\
         Stable-Diffusion-XL & 86.63(3) & 86.01(3) & 85.77(3) & 25.93(2) & 25.81(2) & 25.85(2) & 89.68(3) & 90.04(3) & 90.44(2) \\
         Pixart-Sigma-XL & 87.04(2) & 87.16(2) & 86.72(2) & 25.78(3) &25.71(3) & 25.75(4) & 90.52(2) & 90.53(2) & 89.99(3) \\
         Stable-Diffusion-1.5 & 76.26(4) & 75.79(4) & 77.32(4) & 25.76(4) &25.58(4) & 25.76(3) & 83.23(4) & 82.64(4) & 83.88(4) \\
         \bottomrule
    \end{tabular}
    \caption{Results on MSCOCO using different metrics and different random seeds. The value under metric name is the corresponding random seed used. The value in the brackets is the ranking under the same seed.}
    \label{tab:1}
    \vspace{-10pt}
\end{table*}

\section{Experiment Setup}
For our evaluation, we select four widely-used text-to-image generation models that represent diverse architectures and capabilities: Stable-Diffusion-3 (SD3) \cite{esser2024scaling}, Stable-Diffusion-XL (SD-XL) \cite{podell2023sdxl}, Stable-Diffusion-1.5 (SD1.5) \cite{Rombach_2022_CVPR}, and PixArt-Sigma-XL (Pixart) \cite{chen2024pixartsigma}. This carefully chosen set ensures comprehensive coverage of the current state-of-the-art in diffusion-based text-to-image generation.

For metrics, we employ three widely applicable metrics: CLIPScore \cite{hessel2021clipscore} , VQAScore \cite{Li_2024_CVPR} , and DSGScore \cite{cho2023davidsonian}. For benchmarking, we use MSCOCO \cite{lin2014microsoft}, selecting 1,000 prompts from its validation split to assess general text-to-image generation capabilities following common practice \cite{esser2024scaling}. Additional implementation details are provided in Appendix \ref{sec:app-2}.

\section{Experiment Results and Analysis}

\subsection{Analysis of Robustness}
\paragraph{Robustness to Randomness}
Our investigation first examines evaluation robustness under random seed variations. Table \ref{tab:1} reveals that both CLIPScore and DSGScore produce inconsistent model rankings across different random seeds, demonstrating their failure to maintain robust evaluation outcomes.

Notably, this inconsistency cannot be solely attributed to narrow score margins between models. For instance, CLIPScore with seed $3407$ shows a $0.10$ performance gap between Pixart and SD-XL, compared to a larger $0.13$ gap between Pixart and SD-1.5. Despite this greater margin, CLIPScore inconsistently ranks Pixart versus SD-1.5 while maintaining stable rankings for SD-XL.

We emphasize that our analysis does not presuppose the rankings produced by VQAScore are inherently ``correct." Rather, our core argument establishes that any metric failing to maintain consistent rankings under random seed variations cannot be considered trustworthy, as there exists no reliable basis to determine which ranking might be "correct" when results fluctuate.

Furthermore, while 1000 prompts constitute a large-scale evaluation in text-to-image research, our findings reveal significant robustness failures even at this scale. This persistent inconsistency underscores fundamental challenges in current evaluation practices.

\textbf{Takeaway 1: An evaluation failing to provide robust ranking under randomness should be viewed less trustworthy, like CLIPScore and DSGScore.}

\paragraph{Robustness to Image Perturbation}
For our image perturbation analysis, we utilize SD-3 (selected for its superior generation performance). The evaluation employs a fixed random seed of 42, with complete results presented in Table \ref{tab:2}.

\begin{table}[htbp]
    \small
    \centering
    \begin{tabular}{c|cc}
    \toprule
         &  Avg. $\Delta J_{i}$ & Max. $\Delta J_{i}$  \\
        \midrule
       VQAScore  & 0.44 & 10.36 \\
       CLIPScore & 0.74 & 7.30 \\
       DSGScore & 3.09 & 50.00\\
       \bottomrule
    \end{tabular}
    \caption{The average and maximum performance gap between original image and perturbated image of different metrics.}
    \label{tab:2}
    \vspace{-10pt}
\end{table}


The results reveal that even a minimal perturbation of 
$1$ pixel value creates significant performance variations. While VQAScore demonstrates relatively better robustness with an average performance gap of $0.44$, both CLIPScore and DSGScore exhibit unacceptably large variations, indicating fundamental robustness limitations.

More concerning are the worst-case scenarios, where this simple perturbation produces dramatic performance gaps across all metrics. Figure \ref{fig:intro} illustrates one such failure case for VQAScore. These extreme cases are particularly problematic as they not only occur unpredictably, but also reveal potential for metric exploitation. Further, this worse case may confuse model development when visually identical inputs produce substantially different evaluations.

\textbf{Takeaway 2: All three metrics fail to maintain a worst case robustness under a simple slight perturbation of image. CLIPScore and DSGScore even fail on average case, revealing a worrying fact of the trustworthiness of their evaluation result.}

\begin{figure*}
    \centering
    \begin{subfigure}[b]{0.3\textwidth}
        \includegraphics[width=1.0\textwidth]{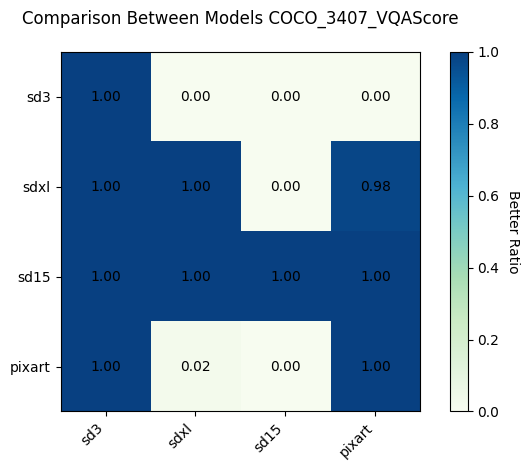}
    \end{subfigure}
    \begin{subfigure}[b]{0.3\textwidth}
        \includegraphics[width=1.0\textwidth]{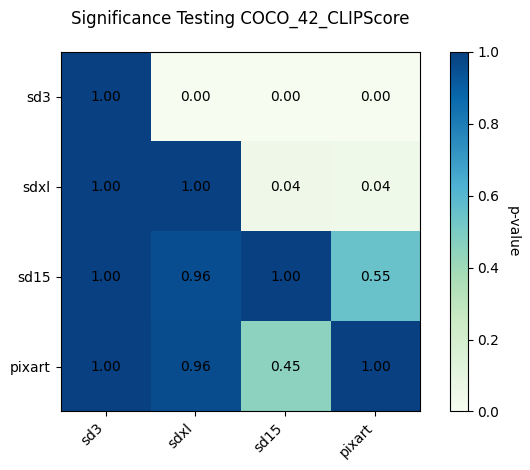}
    \end{subfigure}
    \begin{subfigure}[b]{0.3\textwidth}
        \includegraphics[width=1.0\textwidth]{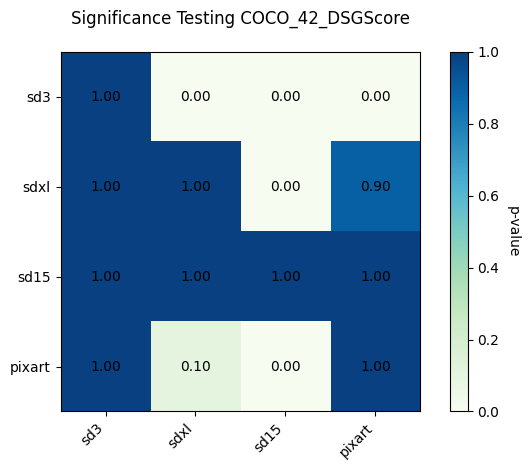}
    \end{subfigure}
    \begin{subfigure}[b]{0.3\textwidth}
        \includegraphics[width=1.0\textwidth]{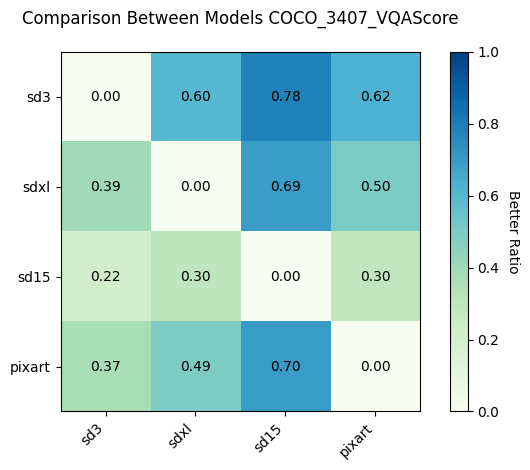}
    \end{subfigure}
    \begin{subfigure}[b]{0.3\textwidth}
        \includegraphics[width=1.0\textwidth]{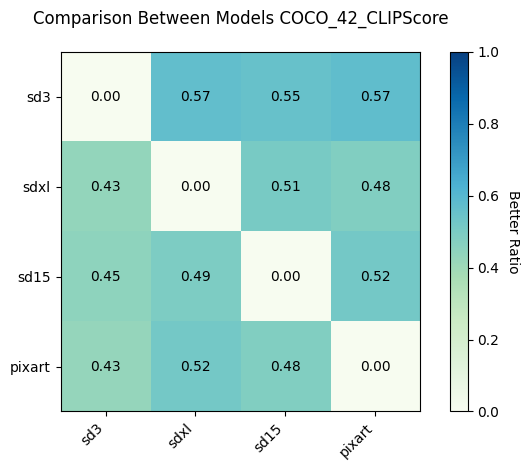}
    \end{subfigure}
    \begin{subfigure}[b]{0.3\textwidth}
        \includegraphics[width=1.0\textwidth]{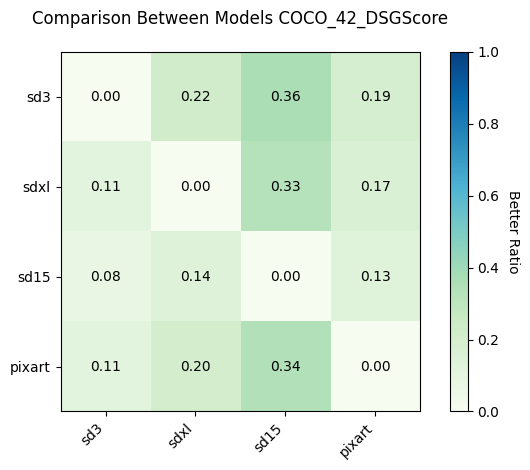}
    \end{subfigure}
    \caption{The T-test p-value and better ratio $R$ between models. The value at $i$-th row and $j$-th column in a matrix represents the result of model $i$ compared against model $j$. The random seed used is shown in the title of the corresponding heatmap.}
    \label{fig:sig}
    \vspace{-10pt}
\end{figure*}

\subsection{Analysis of Significance}
We present the p-value of paired T-test and dominance ratio $R$ to explore the significance of the comparison using different metrics in Figure \ref{fig:sig}. 

We first examine the statistical significance of model comparisons. Following conventional standards, we consider results with p-value $<0.05$ as statistically significant. Our analysis using VQAScore reveals several notable findings: SD-3 demonstrates statistically significant superiority over SD-XL, Pixart, and SD-1.5. More surprisingly, Pixart shows significant improvement over SD-XL (87.16 vs 86.01) in Table \ref{tab:1}, seed 3407), indicating that even 
a score difference of $1$ can reflect meaningful performance gaps between models.

This phenomenon is not unique to VQAScore. CLIPScore exhibits a similar pattern, showing SD-XL significantly superior to Pixart (25.93 vs 25.78). Similarly, DSGScore suggests potential superiority of Pixart over SD-XL ($p=0.10$), despite their small score difference (90.52 vs 
89.68). These consistent findings across metrics challenge conventional assumptions, suggesting that statistically significant improvements may occur with smaller metric differences than previously believed. So are current standards for determining meaningful model improvements  excessively stringent?

To further investigate this phenomenon, we examine the dominance ratio - the probability that model $i$ generates superior results to model $j$ for a given prompt. We reach another shocking observation that, even if model $i$ is significantly better than model $j$, the generation result of model $i$ may not bear a large probability of being better than model $j$. Considering the significance derived using VQAScore as mentioned before, actually SD-XL bears a 50\% probability generating a ``better" result, while Pixart bears $49\%$, even less than SD-XL! Further, SD-3 just bears $60\%$ probability of generating better results, while SD-XL still bears $40\%$ probability of generating better results, even with a VQAScore gap of $4.5$. 

In this context, the problem is not about metrics only. The problem is, how we view ``significance". If we simply view ``significance" as a statistical test,  we can happily accept a small improvement in metrics since it is enough to demonstrate this significance. However, if we would like to guarantee a better generation performance, it is still essential to seek an even larger metric improvement.

\textbf{Takeaway 3: A small difference between metrics is enough to reveal a ``significance" in statistical analysis. However, this ``significance" may not be directly interpreted to actual model performance, while a trustworthy evaluation should take both aspects into account.}

\section{Conclusion}
In this work, we introduce two important aspects, Robustness and Significance, that evaluation frameworks should focus. We conduct a wide range of experiments and reveal some important conclusions that should be taken into consideration when conducting future evaluation on image-text alignment.

\section*{Limitations}

The main limitation of this work is that it does not provide a better evaluation framework to address the problems discussed in this paper. A better evaluation framework is left for future works.

\bibliography{main}

\appendix

\section{Related Works}
\label{sec:appendix}
There are many metrics and benchmarks focusing on evaluating image-text alignment in text-to-image generation.

CLIP-Score \cite{radford2021learning, hessel2021clipscore} evaluates image-text alignment by computing the cosine similarity of CLIP embeddings. VQAScore \citep{lin2024evaluating} queries a VQA model to determine if the image corresponds to the prompt, using the "Yes" logit as the metric. T2I-CompBench \citep{t2icompbench} leverages MiniGPT-4 \citep{zhu2023minigpt} CoT to generate an alignment score. 

Decomposition-based metrics break down text prompts into smaller components and assess the accuracy of each part. TIFA \citep{Hu_2023_ICCV} generates visual questions and uses a VQA model to verify the correctness of each component. T2I-CompBench \citep{t2icompbench} employs Blip-VQA \citep{li2022blip}, while DavidSceneGraph \citep{cho2023davidsonian} and VQ$^{2}$ \citep{yarom2024you} are similar to TIFA. MHalu-Bench \citep{chen2024unified} builds a pipeline of tools to check the correctness of each component directly. ConceptMix\cite{wu2024conceptmix} is a more complicated benchmark.

Some other benchmarks includes \cite{gokhale2023benchmarkingspatialrelation, patel2024conceptbed}, GenEval \citep{GenEval}.

However, all these works consider only alignment with human annotation to evaluate the validity of their evaluation. \cite{gecko} takes a step forward and points out different human evaluation template influences evaluation results, but it still focuses only on human annotation. To the best of our knowledge, we are the first to explore image-text alignment evaluation from the inner properties of the evaluation. 

\section{Experiment Details}
\label{sec:app-2}

We use the default inference hyper-parameter of each model used. We list the details as follows:

\begin{table}[htbp]
    \centering
    \setlength{\tabcolsep}{1.2mm}
    \begin{tabular}{c|c|c}
        \toprule
        Model Name  & $T$ & $w$ \\
        \midrule
        Stable-Diffusion-3 & 28 & 7.0 \\
        Stable-Diffusion-XL & 50 & 5.0 \\
        Stable-Diffusion-1.5 & 50 & 7.5 \\
        PixArt-$\Sigma$-XL & 20 & 4.5 \\
        \bottomrule
    \end{tabular}
    \caption{Details of our inference hyper-parameter. $T$ represents total denoising steps and $w$ represents guidance scale.}
    \label{tab:parameter}
\end{table}

\end{document}